\documentclass[journal,twoside]{ieeecolor}

\usepackage{arxiv}
\usepackage{cite}
\usepackage{amsmath,amssymb,amsfonts}
\usepackage{algorithmic}
\usepackage{graphicx}
\usepackage{textcomp}
\usepackage{bbm} 
\usepackage{booktabs}
\usepackage{multirow}
\usepackage{pifont}
\usepackage[hidelinks]{hyperref}

\usepackage[table,dvipsnames]{xcolor}

\def\BibTeX{{\rm B\kern-.05em{\sc i\kern-.025em b}\kern-.08em
    T\kern-.1667em\lower.7ex\hbox{E}\kern-.125emX}}
\makeatletter
\def\ps@titlepagestyle{%
  \def\@oddhead{\hfill\thepage}%
  \def\@evenhead{\thepage\hfill}%
  \let\@oddfoot\@empty
  \let\@evenfoot\@empty
}
\makeatother
\begin{document}
\title{SurgLAT: Surgical Latent Attention Tracking for Depth-Aware Robotic Laparoscope Control}

\author{%
Rulin Zhou,
Qiujie Song,
Yujie Ma,
An Wang,
Wanhao Liu,
Guoheng Ma,
Yidu Wang,
Guankun Wang,
Xingrong Diao,
Jiankun Wang,
Chaowei Zhu,
Xianming Liu,
and Hongliang Ren
\thanks{R. Zhou, Q. Song, A. Wang, W Liu, Y. Wang, G. Wang, and H. Ren are with the Department of Electronic Engineering, The Chinese University of Hong Kong, Hong Kong SAR 999077, China and Shenzhen Loop Area Institute, Shen Zhen 518055, China  (email: zhourulin@link.cuhk.edu.hk; hlren@ee.cuhk.edu.hk).}
\thanks{Y. Ma is with the College of Mechatronics and Engineering, Shenzhen University, Shenzhen 518060, China.}
\thanks{G. Ma, X. Diao, and J. Wang are with the Department of Electronic and Electrical Engineering, Southern University of Science and Technology, Shenzhen 518055, China.}
\thanks{C. Zhu and X. Liu are with the Shenzhen People’s Hospital, Shenzhen 518020, China.}
}

\maketitle

\begin{abstract}
Autonomous laparoscopic camera control requires continuous understanding of the surgeon’s operative intent in dynamic surgical scenes, where the target operative region is not a stable physical object but a latent and temporally evolving attention state. In this work, we present Surgical Latent Attention Tracking (\textbf{SurgLAT}), a causal online framework for latent surgical attention modeling and autonomous laparoscopic view control. SurgLAT uses a frozen DINOv3 encoder and a state-conditioned spatial token mixer to extract operative evidence under a memory-guided spatial prior, while a selective causal latent memory module jointly models short-term motion continuity and long-horizon surgical intent evolution through dynamic retrieval of current, recent, and historical latent states. The learned latent surgical attention state is decoded into a probabilistic attention heatmap and operative region for downstream endoscope guidance. Beyond perception, we further introduce a robotic deployment framework with explicit laparoscopic Remote Center of Motion (RCM) constrained control based on virtual-axis formulation, together with redundancy-aware null-space initialization for stable and smooth manipulator motion. We validate the full system on real laparoscopic surgical videos and a physical robotic laparoscope platform. Experimental results demonstrate robust online operative-region tracking and stable autonomous endoscopy adjustment under occlusion, rapid motion, and target transitions, highlighting the effectiveness of latent surgical intent modeling for surgical autonomy. Code and video demo are available at~\url{https://surglat-home-page.pages.dev/}
\end{abstract}


\section{Introduction}
\label{sec:introduction}

\begin{figure*}[t]
\centering
\centerline{\includegraphics[width=\linewidth]{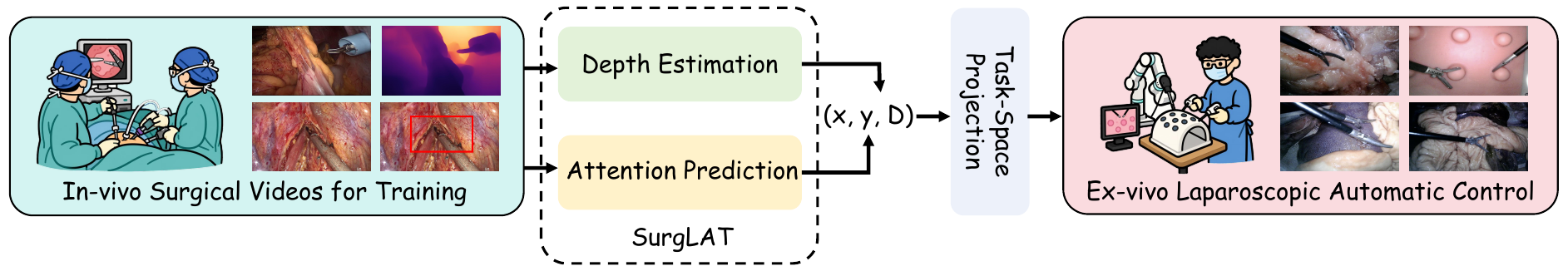}}
    \caption{
    Overview of the proposed perception-to-control pipeline: in-vivo surgical videos are used to learn surgical attention and depth-aware scale estimation, which are converted into task-space targets $(x,y,D)$ for ex-vivo robotic laparoscopic automatic control. }
    \label{fig:introduction}
    \vskip -0.15in
\end{figure*}

\IEEEPARstart{M}{inimally} invasive surgery (MIS) has become a standard paradigm in modern surgical practice due to its reduced trauma, faster recovery, and improved patient outcomes compared with open surgery~\cite{haidegger2022robot, dupont2021decade}. In laparoscopic surgery, the endoscopic video stream is the surgeon's primary perceptual interface to the operative field, making a stable and clinically meaningful field of view (FoV) essential for safe manipulation and efficient surgical workflow. However, conventional laparoscope positioning is commonly performed by an assistant following verbal instructions, which is susceptible to communication latency, fatigue, hand tremor, and inconsistent viewpoint adjustment~\cite{merola2002comparison, wagner2021learning}. These limitations may cause image instability, loss of critical anatomy, and interruptions to the surgical flow, motivating increasing interest in autonomous robotic laparoscope control.

Existing FoV control methods mainly rely on explicit human commands or automatic image-guided strategies. Voice-, gaze-, head-motion-, or foot-pedal-based interfaces allow direct camera regulation, but still impose additional cognitive and operational burden on the surgeon~\cite{fujii2013gaze}. Automatic approaches typically define the desired camera target using hand-crafted visual proxies, such as tool centroids, instrument distributions, or visual-servoing objectives~\cite{osa2010framework, yang2019adaptive, huang2022surgeon}. Although effective for basic tool-following scenarios, these geometric objectives do not directly capture the task-dependent operative focus, which may correspond to a tool--tissue interaction, anatomical boundary, suturing region, or inspection area. Recent learning-based studies have attempted to infer intent-aware viewpoints, and SurgAtt-Tracker~\cite{zhou2026surgatttrackeronlinesurgicalattention} further formulates online surgical attention tracking with expert-annotated operative attention regions. Nevertheless, existing attention-guided methods are still largely limited to 2D localization or image-plane recentering, without explicitly modeling surgical attention as a causal latent state or converting it into physically feasible robotic laparoscope motion.

Learning surgical attention for autonomous laparoscope control remains challenging. Surgical attention is not an explicit object category, but a latent operative intent shaped by surgical phase, tool--tissue interaction, and anatomical context. Meanwhile, laparoscopic videos are often affected by smoke, specular highlights, bleeding, tissue deformation, instrument occlusion, and rapid camera motion, which makes frame-wise prediction unstable. The operative focus also evolves with both temporal continuity and abrupt transitions: it should remain stable during continuous manipulation, yet adapt quickly when the surgeon changes instruments or shifts the target region. In addition, autonomous laparoscope control requires accurate and temporally stable visual targets that can be converted into physically feasible robot motion under the trocar-centered remote-center-of-motion (RCM) constraint.

To address these challenges, we introduce Surgical Latent Attention Tracking (\textbf{SurgLAT}), a perception-to-control framework for autonomous laparoscopic FoV control. Given streaming endoscopic video, SurgLAT models surgical attention as a causal latent operative state rather than an instrument-following point or a generic object box. This latent state is decoded into a probabilistic attention heatmap, from which an attention center and an attention-centered operative region are derived for image-plane FoV guidance. In parallel, a depth-aware branch aggregates relative depth responses within the predicted attention region to estimate the operative scale, enabling adaptive zoom-in/out regulation along the viewing direction.

For robotic deployment, the predicted attention target and depth-aware scale cue are converted into task-space commands $(x,y,D)$, where $(x,y)$ guides lateral recentering and $D$ regulates depth-wise camera motion. These commands are executed by a virtual-axis-based RCM-constrained controller, while a redundancy-aware null-space initialization strategy improves the feasible workspace and smoothness of the 7-DoF manipulator. In this way, SurgLAT integrates latent attention prediction, depth-aware FoV regulation, and physically constrained robotic execution into a unified closed-loop laparoscope control framework.

The main contributions are summarized as follows:
(i) We design a \textbf{causal latent surgical attention model} that estimates operative focus from streaming laparoscopic video through memory-conditioned spatial reasoning and selective temporal updating.
(ii) We introduce a \textbf{depth-aware operative scale estimation branch} that converts monocular relative depth responses into an adaptive cue for zoom-in/out laparoscope regulation.
(iii) We develop a \textbf{virtual-axis-based RCM-constrained robotic control strategy} with redundancy-aware null-space initialization for real-time FoV adjustment on a physical 7-DoF laparoscope platform.
(iv) Extensive experiments on \textbf{SurgAtt-1.16M and real-world robotic validation} demonstrate accurate surgical attention prediction and stable closed-loop FoV control under occlusion, multi-instrument interference, and dynamic tissue interaction.

\section{Related Work}

\subsection{Surgical Attention Modeling}

\subsubsection{Attention-Aware Field-of-View Guidance}

Autonomous laparoscope control requires a reliable visual target that reflects the surgeon's operative intent. Early image-guided methods commonly define this target using hand-crafted geometric cues, such as a single-tool centroid, a multi-tool center, or an enclosing region around visible instruments~\cite{osa2010framework, yang2019adaptive}. These methods are simple and effective for basic tool-following scenarios, but they implicitly assume that the desired field of view (FoV) is determined by instrument geometry. In real procedures, however, the clinically informative region may correspond to a tissue--tool interaction, an exposed anatomical boundary, a suturing site, or an inspection area rather than the geometric center of visible tools. To move beyond purely geometric objectives, recent studies have explored intent-aware FoV guidance using gaze, voice commands, surgeon preference modeling, imitation learning, and action-aware prediction~\cite{fujii2013gaze, sandoval2021towards, li20223d}. These methods provide more semantically meaningful camera targets, but many of them still rely on explicit user input, platform-specific action prediction, or predefined control policies. SurgAtt-Tracker~\cite{zhou2026surgatttrackeronlinesurgicalattention} formulates surgical FoV guidance as online surgical attention tracking and introduces expert-annotated operative attention regions. 


\subsubsection{Temporal Robustness in Surgical Video Prediction}

Surgical attention estimation is inherently temporal, since the operative focus should remain stable during continuous manipulation while adapting to instrument changes, target shifts, and unexpected events. Prior studies have improved the stability of image-guided laparoscope control through temporal smoothing, probabilistic modeling, and predictive tracking. Visual-servoing-based systems regularize target trajectories or camera motion to suppress high-frequency jitter~\cite{gruijthuijsen2022robotic}, probabilistic frameworks encode surgeon preferences or instrument-state distributions from historical observations~\cite{li2021data, li2024gmm}, and hybrid tracking modules combine online tracking with future-position prediction to compensate for latency~\cite{iovene2024hybrid}. However, these methods mainly stabilize explicit visual proxies, such as tools, centroids, or selected geometric points. In contrast, surgical attention is a task-dependent latent state that may shift between instruments, tissues, anatomical structures, and interaction regions, motivating a causal temporal model that preserves attention continuity while allowing rapid state updates~\cite{wang2026visionsafeenhanced}.


\subsection{RCM-Constrained Laparoscope Control}

Autonomous laparoscope control must satisfy the remote-center-of-motion (RCM) constraint imposed by the trocar port, which restricts the laparoscope shaft to rotate around the incision point and is essential for safe minimally invasive manipulation. Prior studies on image-based visual servoing and model-based robotic camera control typically define image-plane errors using tool tips, instrument centroids, or selected visual features, and then compute camera motion to minimize the error~\cite{osa2010framework, ma2020visual, zhang2023visual}. Other methods incorporate camera-quality assessment, heuristic policies, constrained optimization, RCM-aware planning, or redundant manipulator control to improve view stability and deployment feasibility~\cite{bihlmaier2014automated, gruijthuijsen2022robotic, li2024gmm, sun2020visual, fozilov2023endoscope, jiang2025robust, pasini2023grace}. However, these approaches are often limited to 2D recentering and remain strongly coupled with instrument geometry. Consequently, the camera may satisfy the geometric tracking objective while failing to maintain the most informative operative view.

Comprehensive integration of learned surgical attention prediction, depth-aware FoV regulation, RCM-constrained execution, and real-world robotic validation remains insufficiently explored. Different from prior work that treats perception and control as loosely connected components, our framework converts the predicted attention center and depth-aware operative scale cue into task-space commands for lateral recentering and zoom-in/out regulation. These commands are executed through a virtual-axis-based RCM-constrained controller with redundancy-aware null-space initialization, enabling physically feasible and intention-aware laparoscope motion.

\begin{figure*}[t]
\centering
\centerline{\includegraphics[width=\linewidth]{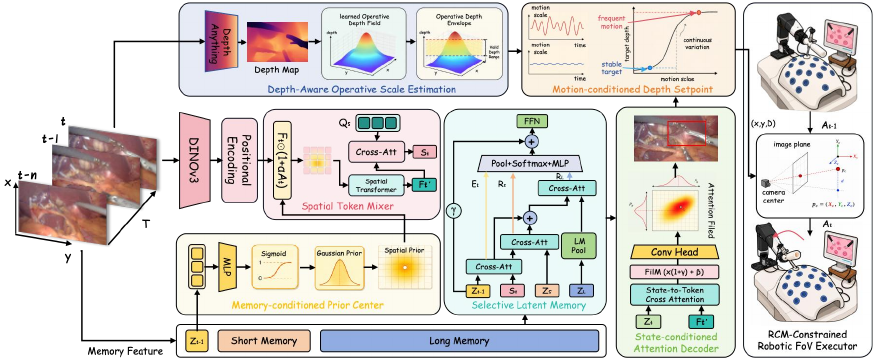}}
    \caption{
   SurgLAT predicts the image-plane operative target $(x,y)$ from streaming laparoscopic video through memory-conditioned spatial reasoning and selective latent memory. In parallel, a depth-aware branch estimates the operative depth envelope, from which a motion-conditioned depth-domain regulation policy derives the viewing-direction depth target $D$ for adaptive zoom-in/out control. The combined task-space target $(x,y,D)$ is executed by an RCM-constrained robotic FoV executor for closed-loop laparoscopic viewpoint adjustment. }
    \label{fig:method}
    \vskip -0.2in
\end{figure*}

\section{Methodology}
\label{sec:Method}

\subsection{Problem Formulation and Overview}

Given a streaming laparoscopic video $\{I_1,\ldots,I_t\}$, our goal is to estimate an operative attention target that indicates where the laparoscope should look and convert it into physically feasible camera motion. Unlike a
persistent object with stable appearance and boundaries, surgical attention is a context-dependent and temporally evolving operative focus. We therefore model it as a causal latent state:
\begin{equation}
(H_t,\hat{B}_t,Z_t)
=
\Phi_{\theta}
(I_t,Z_{t-1},\mathcal{Z}_s,\mathcal{Z}_l),
\label{eq:problem_formulation}
\end{equation}
where $H_t$, $\hat{B}_t$, and $Z_t$ denote the attention field, operative region, and updated latent state, respectively. SurgLAT selectively integrates current visual evidence, short-term motion continuity, and long-term surgical context. The predicted attention region provides image-plane guidance, while a depth-aware scale cue is converted into executable FoV adjustment by the downstream RCM-constrained controller.

\subsection{Latent Surgical Attention Tracking Model}

\subsubsection{Visual Token Encoding}
For each incoming laparoscopic frame $I_t$, we first extract dense visual tokens using a frozen DINOv3~\cite{simeoni2025dinov3} visual foundation encoder. The encoder maps the current frame into a patch-level feature grid $F_t$, where each token encodes local semantic and appearance evidence from the endoscopic scene. Since the model is designed for causal online inference, this encoding is performed independently for each frame and does not use future observations or bidirectional temporal context. We then project the DINOv3 features into a unified hidden dimension and add two-dimensional positional embeddings, producing spatially aware visual tokens for subsequent attention modeling.

\subsubsection{Memory-Conditioned Spatial Token Mixer}

In laparoscopic surgery, the operative attention region is usually a spatially concentrated focus rather than a uniformly distributed scene cue. Although DINOv3 provides dense and semantically rich visual tokens over the entire laparoscopic scene, these tokens are not explicitly biased toward the current operative focus. Meanwhile, the attention region in the current frame is often correlated with its previous location, as tool--tissue interaction and camera motion evolve continuously over time. Therefore, instead of treating each frame as an independent global search problem, SurgLAT uses the previous latent attention state $Z_{t-1}$ to generate a memory-conditioned spatial prior for the current frame. This prior provides a coarse reference of where the operative focus is likely to appear, enabling more focused feature refinement while preserving global contextual reasoning.

As shown by the \textbf{Memory-conditioned Prior Center} and \textbf{Spatial Token Mixer} in Fig.~\ref{fig:method}, the previous latent state tokens $Z_{t-1}$ are first used to predict a normalized prior center $p_t$, which is then converted into a Gaussian spatial prior $A_t$ over the DINO token grid. Instead of cropping or masking the image, the prior softly modulates the current visual tokens:
\begin{equation}
F_t^{\mathrm{prior}} = F_t \odot (1 + \alpha A_t),
\end{equation}
where $F_t$ denotes the projected DINOv3 token grid with 2D positional encoding, and $\alpha$ controls the prior strength. The modulation enhances tokens near the expected operative focus without suppressing the remaining scene context. Therefore, the model retains global visual evidence and can recover when the memory-conditioned prior is inaccurate.

The modulated token grid is then processed by a spatial transformer to obtain refined dense visual tokens.
\begin{equation}
F'_t = \mathrm{SpatialTransformer}(F_t^{\mathrm{prior}}).
\end{equation}
The spatial transformer maintains two token streams: dense image tokens and a small set of learnable spatial evidence tokens. The dense tokens perform self-attention over the full spatial grid to exchange global context while preserving spatial layout. The learnable spatial tokens then serve as task-specific evidence queries and cross-attend to the refined dense tokens:
\begin{equation}
S_t = \mathrm{CrossAttn}(Q_s, F'_t, F'_t),
\end{equation}
where $Q_s$ denotes the learnable spatial queries. These spatial evidence tokens are further refined through self-attention and feed-forward layers, allowing interactions among multiple evidence slots. Finally, the refined dense grid $F'_t$ is passed to the state-conditioned heatmap decoder, while the compact spatial evidence tokens $S_t$ are sent to the selective latent memory module for causal state updating.

\subsubsection{Source-Selective Causal Latent Memory}

Surgical attention should remain temporally stable during continuous manipulation while rapidly adapting when the operative focus changes. To capture these complementary properties, SurgLAT explicitly decomposes the state update into current visual evidence, short-term continuity, and long-term surgical context.

As shown in Fig.~\ref{fig:method}, the previous latent state $Z_{t-1}$ queries the current spatial tokens $S_t$ to obtain the current evidence $E_t$. The current evidence then retrieves a short-term residual $R_t^{s}$ from the recent-state cache $\mathcal{Z}_s$. Meanwhile, the long-term cache $\mathcal{Z}_l$ is compressed into memory prototypes $P_t^{l}=\operatorname{LMPool}(\mathcal{Z}_l)$, which are queried by $E_t+R_t^{s}$ to obtain the long-term residual $R_t^{l}$.

Rather than combining these sources with fixed weights, a source router predicts their frame-specific contributions:
\begin{equation}
[w_t^{c},w_t^{s},w_t^{l}]
=
\operatorname{Softmax}
\left(
\operatorname{MLP}
([\bar{E}_t,\bar{R}_t^{s},\bar{R}_t^{l}])
\right),
\label{eq:source_router}
\end{equation}
where the barred variables denote pooled source representations.
The source-selective residual is
\begin{equation}
\Delta Z_t
=
w_t^{c}E_t
+
w_t^{s}R_t^{s}
+
w_t^{l}R_t^{l}.
\label{eq:source_selective_residual}
\end{equation}
The latent attention state is updated as
\begin{equation}
\hat{Z}_t
=
Z_{t-1}
+
\gamma\Delta Z_t,
\qquad
Z_t
=
\hat{Z}_t
+
\operatorname{FFN}(\hat{Z}_t),
\label{eq:latent_state_update}
\end{equation}
where $\gamma$ is a learnable residual scale initialized to $0.1$. This source-selective update preserves short-term continuity during stable manipulation while increasing current or long-term evidence when the operative focus changes.

\subsubsection{State-Conditioned Heatmap and ROI Decoding}

After the source-selective state update, SurgLAT projects the temporally accumulated latent state back to the dense spatial tokens and decodes an attention heatmap and operative ROI. Since surgical attention corresponds to a functional operative focus rather than an object with explicit boundaries, directly regressing all box coordinates $(x,y,w,h)$ from a global feature can be unstable. We therefore decouple center localization from scale estimation: the attention center is inferred from a spatial heatmap, while a lightweight scale head predicts the ROI extent. As shown in Fig.~\ref{fig:method}, given the refined dense tokens $F'_t$ and the updated latent state tokens $Z_t$, the decoder first conditions the visual grid on the current latent state using state-to-token cross-attention and FiLM-style channel modulation. The conditioned dense grid is then passed to a convolutional head to predict the attention heatmap $H_t$. This heatmap formulation preserves spatial layout and provides a smoother alternative to unconstrained coordinate regression.

To obtain a differentiable attention center, we marginalize the heatmap into horizontal and vertical distributions:
\begin{equation}
P_x = \mathrm{Softmax}(\mathrm{LogSumExp}_y(H_t)),
\end{equation}
\begin{equation}
P_y = \mathrm{Softmax}(\mathrm{LogSumExp}_x(H_t)).
\end{equation}
The center is then computed as the expectation of the two axis-wise distributions:
\begin{equation}
\hat{c}_t =
(\mathbb{E}_{P_x}[x],\mathbb{E}_{P_y}[y]),
\end{equation}
which avoids the discontinuity of hard argmax and yields continuous center coordinates from the spatial response.

For ROI scale estimation, a lightweight scale head predicts the width and height $\hat{s}_t=(\hat{w}_t,\hat{h}_t)$ conditioned on the latent attention state. The final attention box is constructed by combining the heatmap-derived center and the predicted scale:
\begin{equation}
\hat{B}_t =
[\hat{x}_t-\hat{w}_t/2,\hat{y}_t-\hat{h}_t/2,
 \hat{x}_t+\hat{w}_t/2,\hat{y}_t+\hat{h}_t/2].
\end{equation}
This design allows the heatmap decoder to focus on stable operative-center localization, while the scale head adaptively estimates the spatial extent of the attention region.

\subsubsection{Loss Function and Training Protocol}

We train SurgLAT with heatmap-assisted box supervision. Since the attention center is the primary variable for laparoscope recentering, while the ROI extent is adaptively estimated by the scale head, the objective jointly supervises the attention heatmap, decoded center, predicted scale, and temporal center displacement. The overall training objective is defined as:
\begin{equation}
\mathcal{L}
=
\mathcal{L}_{\mathrm{hm}}
+
\mathcal{L}_{\mathrm{center}}
+
\mathcal{L}_{\mathrm{scale}}
+
\mathcal{L}_{\mathrm{temp}} .
\end{equation}

The heatmap loss supervises the predicted attention field using a Gaussian target centered at the ground-truth center:
\begin{equation}
\mathcal{L}_{\mathrm{hm}}
=
\mathrm{BCEWithLogits}(H_t, G(c_t^{*})),
\end{equation}
where $H_t$ denotes the predicted heatmap logits, $c_t^{*}$ is the ground-truth center, and $G(c_t^{*})$ is the corresponding Gaussian heatmap. The center loss further constrains both the axis-wise center distributions and the continuous expected center:
\begin{equation}
\mathcal{L}_{\mathrm{center}}
=
\mathcal{L}_{\mathrm{axis}}
+
\mathrm{SmoothL1}(\hat{c}_t,c_t^{*}),
\end{equation}
where $\mathcal{L}_{\mathrm{axis}}$ is the cross-entropy loss over horizontal and vertical center bins, and $\hat{c}_t$ is the expected center decoded from the axis distributions.

To supervise the adaptive ROI extent, we apply a scale regression loss to the width and height predicted:
\begin{equation}
\mathcal{L}_{\mathrm{scale}}
=
\mathrm{SmoothL1}
\left(
\log(\hat{s}_t+\epsilon),
\log(s_t^{*}+\epsilon)
\right),
\end{equation}
where $\hat{s}_t=(\hat{w}_t,\hat{h}_t)$ and $s_t^{*}=(w_t^{*},h_t^{*})$ denote the predicted and ground-truth ROI scales, respectively. The logarithmic form reduces sensitivity to absolute box size and stabilizes scale learning.

To encourage temporally coherent predictions, we regularize the predicted inter-frame center displacement:
\begin{equation}
\mathcal{L}_{\mathrm{temp}}
=
\frac{1}{T-1}
\sum_{t=2}^{T}
\mathrm{SmoothL1}
\left(
\hat{c}_t-\hat{c}_{t-1},
c_t^{*}-c_{t-1}^{*}
\right).
\end{equation}
This term penalizes inconsistent motion while allowing the prediction to follow the ground-truth attention trajectory. It improves temporal stability without altering the causal inference setting, since the model still predicts each frame using only the current observation and past latent states.

We adopt a two-stage training protocol to balance optimization stability and online consistency. In the first stage, SurgLAT is trained on short video clips to stabilize the spatial reader, latent memory update, state-conditioned heatmap decoder, and scale head on top of frozen DINO features. In the second stage, we switch to an online streaming protocol that matches deployment. Consecutive segments from the same video are processed in temporal order, and the latent memory is carried across segment boundaries instead of being reset at each clip. This streaming protocol aligns training with causal inference and enables stable memory propagation during continuous video prediction.

\subsection{Depth-Aware Operative Scale Regulation}

\subsubsection{Attention-Weighted Operative Depth Estimation}

The predicted attention center $(\hat{x}_t,\hat{y}_t)$ specifies the lateral recentering target, but does not determine the appropriate axial viewing scale. We therefore estimate the relative operative depth within the predicted attention region using a pretrained
monocular depth model. Since monocular depth is scale-ambiguous, the predicted depth map $D_t$ is normalized within each frame by robust percentile normalization and used only as a relative geometric cue rather than a metric camera-to-tissue distance. The depth orientation
is standardized such that larger normalized values indicate farther operative surfaces. Given the predicted heatmap $H_t$, we compute an attention-weighted operative depth score:
\begin{equation}
d_t^{\mathrm{op}}
=
\frac{
\sum_{i,j}
\alpha_t(i,j)\tilde{D}_t(i,j)
}{
\sum_{i,j}
\alpha_t(i,j)+\epsilon
},
\qquad
\alpha_t(i,j)=\sigma(H_t(i,j)),
\label{eq:operative_depth}
\end{equation}
where $\tilde{D}_t$ denotes the normalized relative depth map. This aggregation focuses the depth estimate on the predicted operative region rather than the full image. From the training set, we estimate a valid relative-depth envelope $\mathcal{E}=[d^{\mathrm{near}},d^{\mathrm{far}}]$, representing the typical viewing-scale range associated with stable laparoscopic observation.

\subsubsection{Motion-Conditioned Depth Setpoint}

The desired axial viewing scale depends on the recent motion of the attention target. A stable target permits a closer view for fine manipulation, whereas frequent target motion benefits from a wider FoV. We therefore compute a normalized motion scale from the recent
attention trajectory:
\begin{equation}
m_t
=
\frac{1}{K d_{\mathrm{img}}}
\sum_{k=1}^{K}
\left\|
\hat{c}_{t-k+1}
-
\hat{c}_{t-k}
\right\|_2,
\label{eq:attention_motion_scale}
\end{equation}
where $K$ is the temporal window length and $d_{\mathrm{img}}$ is the image diagonal. The motion scale is mapped to a depth interpolation factor:
\begin{equation}
\gamma(m_t)
=
\operatorname{clip}
\left(
\frac{
m_t-m_{\min}
}{
m_{\max}-m_{\min}+\epsilon
},
0,1
\right).
\label{eq:depth_interpolation_factor}
\end{equation}
The desired depth setpoint is selected within the valid relative-depth envelope:
\begin{equation}
d_t^{\mathrm{set}}
=
\left(1-\gamma(m_t)\right)d^{\mathrm{near}}
+
\gamma(m_t)d^{\mathrm{far}}.
\label{eq:motion_conditioned_depth_setpoint}
\end{equation}
Thus, low target motion favors a closer operative view, whereas high target motion shifts the setpoint toward a wider FoV for robust tracking and exploration.

\subsubsection{Depth-Domain Axial Regulation}

The estimated operative depth $d_t^{\mathrm{op}}$ describes the current viewing scale, while $d_t^{\mathrm{set}}$ specifies the desired scale. Their difference defines the relative depth-domain error, which is converted into a bounded insertion target:
\begin{equation}
e_t^{d}
=
d_t^{\mathrm{op}}
-
d_t^{\mathrm{set}},
\qquad
\lambda_t^{\mathrm{target}}
=
\operatorname{clip}
\left(
\lambda_{t-1}^{\mathrm{target}}
+
k_d e_t^{d},
\lambda_{\min},
\lambda_{\max}
\right),
\label{eq:depth_domain_regulation}
\end{equation}
where $k_d$ is the axial regulation gain and $[\lambda_{\min},\lambda_{\max}]$ denotes the valid insertion range. The relative-depth signal is therefore used as feedback for axial
view regulation rather than interpreted as an absolute metric depth. The resulting $\lambda_t^{\mathrm{target}}$ is executed by the hierarchical RCM-constrained controller described in Sec.~IV-C.

\section{Robotic Laparoscope Control}
\label{sec:robotic-control}

SurgLAT predicts an image-plane target and an axial depth setpoint for autonomous
laparoscopic viewpoint adjustment. To execute these commands on a physical
robot, the laparoscope must satisfy the Remote Center of Motion (RCM) constraint
imposed by the trocar point $p_{\mathrm{trocar}}$. In addition, the initial
manipulator configuration strongly affects the feasible rotational workspace
during subsequent tracking. We therefore combine RCM-constrained target tracking
with redundancy-aware initial configuration optimization.

\subsection{RCM-Constrained Target Tracking}

\subsubsection{RCM Modeling}

To model insertion motion while maintaining the trocar constraint, we augment
the 7-DoF manipulator configuration $q$ with a virtual insertion coordinate
$\lambda$. Let $p_{\mathrm{ee}}(q)$ and $p_{\mathrm{cam}}(q)$ denote two
points on the laparoscope axis. The RCM point is defined as:
\begin{equation}
p_{\mathrm{rcm}}(q,\lambda)
=
p_{\mathrm{ee}}(q)
+
\lambda
\bigl(
p_{\mathrm{cam}}(q)-p_{\mathrm{ee}}(q)
\bigr).
\end{equation}
Since the laparoscope is rigidly attached to the manipulator, differentiating
$p_{\mathrm{rcm}}$ gives
$
\dot{p}_{\mathrm{rcm}}
=
J_{\mathrm{rcm}}
\begin{bmatrix}
\dot{q}^{T} & \dot{\lambda}
\end{bmatrix}^{T},
$
where
$
J_{\mathrm{rcm}}
=
\begin{bmatrix}
(1-\lambda)J_{p_{\mathrm{ee}}}
+
\lambda J_{p_{\mathrm{cam}}}
&
p_{\mathrm{cam}}-p_{\mathrm{ee}}
\end{bmatrix}.
$
The RCM constraint is enforced by driving $p_{\mathrm{rcm}}$ to the fixed $p_{\mathrm{trocar}}$.

\subsubsection{Target-Oriented Laparoscope Rotation}

The tracking objective is to keep the SurgLAT-predicted target region near the
image center. Given the predicted image target $r_{\mathrm{pix}}$, its
estimated depth, camera intrinsic matrix $K$, and camera pose
$T_{\mathrm{cam}}(q)$, we back-project the target to a world-frame point
$r_w$. Let $p_w$ be the RCM point, $O_w$ the camera optical center, and
$a_w$ the current optical-axis direction. During target tracking, $\lambda$ is kept fixed, so the task reduces to finding an incremental laparoscope rotation $R_{\Delta}=\exp([\omega]_{\times})$. The rotated optical axis is
$a'_w=R_{\Delta}a_w$, and the target direction from the rotated optical center
is
$
b'_w
=
\frac{
r_w-p_w-R_{\Delta}(O_w-p_w)
}{
\left\|
r_w-p_w-R_{\Delta}(O_w-p_w)
\right\|
}.
$
We solve the following alignment objective:
\begin{equation}
\min_{\omega}
\left\|
a'_w \times b'_w
\right\|^2
+
\eta \max(0,-{a'_w}^{T}b'_w),
\end{equation}
where the first term minimizes angular misalignment and the second term avoids
the degenerate case where the optical axis points away from the target. The
optimized rotation defines the desired laparoscope orientation as
$R_{\mathrm{des}}=R_{\Delta}^{*}R_{\mathrm{cur}}$.

\subsection{Initial Configuration Optimization}

Because the laparoscope rotates during tracking, a favorable initial joint configuration can enlarge the rotational workspace. For a fixed initial laparoscope pose, the 7-DoF manipulator has one redundant DoF. Since axial rotation around the laparoscope does not change the viewing direction or violate the RCM constraint, we relax this axial constraint and search in a two-dimensional redundant space.

Let $p_{\mathrm{ee},0}$ and $u_0$ denote the initial end-effector position
and laparoscope-axis direction. The relaxed 5-DoF initial pose constraint is
$
e_{\mathrm{ini}}(q)
=
\begin{bmatrix}
p_{\mathrm{ee}}(q)-p_{\mathrm{ee},0} &
u(q)\times u_0
\end{bmatrix}^{T},
\qquad
J_{\mathrm{ini}}(q)
=
\frac{\partial e_{\mathrm{ini}}}{\partial q}.
$
To evaluate the local rotational capability under the RCM constraint, we project
the axis-direction Jacobian into the RCM null space:
$
J_{\mathrm{rot}}(q)
=
J_u(q)N_{\mathrm{rcm}}^{q}(q),
$
where $J_u(q)$ is the Jacobian of the laparoscope-axis direction and
$N_{\mathrm{rcm}}^{q}$ denotes the null-space of the RCM Jacobian.

The optimal initial configuration is obtained by maximizing the rotation score:
\begin{equation}
\begin{split}
q^*
=
\arg\max_q
\left(
\sigma_{\min}^{+}(J_{\mathrm{rot}})
-
w_{\kappa}\kappa(J_{\mathrm{rot}})
+
w_m m(q)
\right), \\
\mathrm{s.t.}
\quad
e_{\mathrm{ini}}(q)=0,
\qquad
q_{\min}\le q\le q_{\max}.
\end{split}
\label{eq:init_opt}
\end{equation}
Here, $\sigma_{\min}^{+}$ is the smallest non-zero singular value of $J_{\mathrm{rot}}$, encouraging larger local rotational capability; $\kappa(\cdot)$ is the condition number, encouraging isotropic motion; and $m(q)$ is a joint-limit margin that avoids joint limits.

To reduce sensitivity to local optima, we first sample candidate seeds in the
null space of $J_{\mathrm{ini}}(q_0)$. Using two null-space basis
vectors $n_1$ and $n_2$, we generate
$q_{\mathrm{seed}}=q_0+\alpha n_1+\beta n_2$, where
$\alpha,\beta\in[-r,r]$. Each seed is projected back to the initial pose
constraint by iterative correction:
\begin{equation}
q^{k+1}
=
\mathrm{clip}
\left(
q^k
-
J_{\mathrm{ini}}(q^k)^{\#}e_{\mathrm{ini}}(q^k),
q_{\min},
q_{\max}
\right).
\end{equation}
The best projected candidate is then used as the initialization for solving
Eq.~\eqref{eq:init_opt}, yielding the final configuration
$q^*$.

\subsection{Hierarchical Task-Space Control}

Both initialization and online target tracking are executed using a hierarchical
task-space controller. The primary task is solved first; the secondary task
is projected into its null space:
\begin{equation}
\dot{\xi}^{*}
=
J_{\mathrm{pr}}^{\#}v_{\mathrm{pr}}
+
N_{\mathrm{pr}}
\left(
J_{\mathrm{s}}N_{\mathrm{pr}}
\right)^{\#}
\left(
v_{\mathrm{s}}
-
J_{\mathrm{s}}J_{\mathrm{pr}}^{\#}v_{\mathrm{pr}}
\right),
\end{equation}
where $\xi=[q^T,\lambda]^T$,
$N_{\mathrm{pr}}=I-J_{\mathrm{pr}}^{\#}J_{\mathrm{pr}}$,
$v_{\mathrm{pr}}=K_{\mathrm{pr}}e_{\mathrm{pr}}$, and
$v_{\mathrm{s}}=K_{\mathrm{s}}e_{\mathrm{s}}$. During initialization, the primary task enforces the RCM constraint, while the secondary task drives the manipulator toward $q^*$. During online tracking, the primary task jointly enforces the RCM constraint and the desired laparoscope orientation $R_{\mathrm{des}}$, while the secondary task regulates insertion depth using $e_{\mathrm{s}}=\lambda_{\mathrm{target}}-\lambda$. This hierarchy ensures trocar safety is maintained while the robot follows the SurgLAT-predicted target and adaptive depth command.

\section{Experiments}

\subsection{Experimental Setup and Details}


\subsubsection{Dataset and Baselines}
We evaluate SurgLAT on the public SurgAtt-1.16M benchmark~\cite{zhou2026surgatttrackeronlinesurgicalattention}, which contains three subsets: SurgAtt-SZPH, SurgAtt-AutoLaparo, and SurgAtt-Hamlyn. SurgAtt-SZPH includes approximately 1M frames from gastrointestinal surgery and is used as the primary in-domain benchmark. We follow the original annotation protocol, where each frame is annotated with a bounding box indicating the operative attention region. We compare SurgLAT with representative tracking-based and detection-based baselines. Tracking-based methods include AQATrack~\cite{xie2024autoregressive}, ODTrack~\cite{zheng2024odtrack}, SPMTrack-B~\cite{cai2025spmtrack}, LoRAT-B~\cite{lorat}, LoRATv2-B~\cite{linloratv2}, and MCITrack~\cite{kang2025exploring}, which are adapted by treating the operative attention region as a temporally tracked target. Detection-based methods include YOLOv11-S/M~\cite{khanam2024yolov11}, YOLOv12-S/M~\cite{tian2025yolov12}, YOLOv26-S/M~\cite{sapkota2025yolo26}, RT-DETR~\cite{lv2023detrs}, RT-DETRv2~\cite{lv2024rtdetrv2improvedbaselinebagoffreebies}, and SurgAtt-Tracker~\cite{zhou2026surgatttrackeronlinesurgicalattention}, which directly predict the operative attention region from visual observations.

\subsubsection{Evaluation Metrics}

Following SurgAtt~\cite{zhou2026surgatttrackeronlinesurgicalattention}, we use Intersection over Union (IoU) to measure ROI overlap and Mean Center Error (MCE) to measure center localization error in pixels. On SurgAtt-SZPH, we additionally report mAP$_{0.5}$, mAP$_{0.75}$, and mAP$_{0.5:0.95}$ to assess localization quality under different IoU thresholds. Runtime is measured in frames per second (FPS).

\begin{table*}[ht]
\centering
\caption{
Comparison with detection-based and tracking-based baselines on three surgical attention tracking benchmarks.
}
\label{tab:main_results}

\resizebox{\textwidth}{!}{
\begin{tabular}{l|ccccc|cc|cc|c}
\toprule

& \multicolumn{5}{c|}{SurgAtt-SZPH}
& \multicolumn{2}{c|}{SurgAtt-AutoLaparo}
& \multicolumn{2}{c|}{SurgAtt-Hamlyn}
& Runtime \\

\cmidrule(lr){2-6}
\cmidrule(lr){7-8}
\cmidrule(lr){9-10}
\cmidrule(lr){11-11}

Model

& IoU$\uparrow$
& MCE$\downarrow$
& mAP$@{0.5}\uparrow$
& mAP$@{0.75}\uparrow$
& mAP$@{0.5:0.95}\uparrow$
& IoU$\uparrow$
& MCE$\downarrow$
& IoU$\uparrow$
& MCE$\downarrow$
& FPS$\uparrow$ \\

\midrule

\rowcolor{green!15}
\multicolumn{11}{c}{\textit{Detection-based}} \\
\midrule

YOLOv11-S~\cite{khanam2024yolov11}
& 0.470 & 79.45 & 0.372 & 0.040 & 0.111
& 0.283 & 193.93
& 0.311 & 69.24
& 41.5 \\

YOLOv11-M~\cite{khanam2024yolov11}
& 0.471 & 82.00 & 0.370 & 0.040 & 0.112
& 0.285 & 221.66
& 0.239 & 80.71
& 40.0 \\

YOLOv12-S~\cite{tian2025yolov12}
& 0.471 & 80.79 & 0.347 & 0.035 & 0.099
& 0.292 & 203.64
& 0.322 & 67.90
& 39.9 \\

YOLOv12-M~\cite{tian2025yolov12}
& 0.480 & 79.12 & 0.396 & 0.043 & 0.119
& 0.304 & 207.36
& 0.328 & 68.20
& 42.0 \\

YOLOv26-S~\cite{sapkota2025yolo26}
& 0.471 & 84.45 & 0.315 & 0.043 & 0.102
& 0.316 & 211.64
& 0.235 & 81.79
& 43.2 \\

YOLOv26-M~\cite{sapkota2025yolo26}
& 0.466 & 87.89 & 0.328 & 0.042 & 0.106
& 0.313 & 215.88
& 0.323 & 65.57
& 40.3 \\

RT-DETR~\cite{lv2023detrs}
& 0.499 & 77.80 & 0.452 & 0.058 & 0.148
& 0.349 & 162.84
& 0.239 & 52.54
& \textbf{76.7} \\

RT-DETRv2~\cite{lv2024rtdetrv2improvedbaselinebagoffreebies}
& 0.493 & 78.32 & 0.437 & 0.046 & 0.139
& 0.330 & 164.64
& 0.259 & 52.07
& \underline{71.3} \\

\midrule

\rowcolor{purple!18}
\multicolumn{11}{c}{\textit{Object Tracking--based}} \\
\midrule

ODTrack~\cite{zheng2024odtrack}
& 0.413 & 77.20 & 0.304 & 0.026 & 0.084
& 0.320 & 186.30
& 0.328 & 51.27
& 25.0 \\

SPMTrack-B~\cite{cai2025spmtrack}
& 0.486 & 79.51 & 0.524 & 0.079 & 0.178
& 0.403 & 161.02
& 0.406 & 46.27
& 26.7 \\

LoRAT-B~\cite{lorat}
& 0.483 & 78.66 & 0.529 & 0.070 & 0.174
& 0.358 & 178.69
& 0.372 & 50.77
& 49.4 \\

LoRATv2-B~\cite{linloratv2}
& 0.501 & 78.15 & 0.571 & 0.102 & 0.206
& 0.424 & 167.52
& 0.308 & 55.55
& 41.3 \\

AQATrack~\cite{xie2024autoregressive}
& 0.528 & 76.56 & 0.602 & 0.162 & 0.246
& 0.417 & 167.69
& 0.418 & 47.91
& 25.7 \\

MCITrack~\cite{kang2025exploring}
& 0.539
& 74.07
& 0.612
& 0.183
& 0.258
& 0.459 & 154.44
& 0.420 & 46.19
& 17.8 \\

SurgAtt-Tracker~\cite{zhou2026surgatttrackeronlinesurgicalattention}
& \underline{0.566}
& \underline{49.92}
& \underline{0.656}
& \underline{0.220}
& \underline{0.280}
& \underline{0.462}
& \underline{121.12}
& \underline{0.443}
& \underline{42.48}
& {12.4} \\

\midrule

SurgLAT (Ours)
& \textbf{0.604}
& \textbf{41.24}
& \textbf{0.669}
& \textbf{0.268}
& \textbf{0.322}
& \textbf{0.527}
& \textbf{113.75}
& \textbf{0.479}
& \textbf{37.97}
& {34.5} \\
\bottomrule
\end{tabular}
}
\vskip -0.15in
\end{table*}

\subsubsection{Implementation Details}
For fair comparison, all models are trained for 10 epochs on an NVIDIA H100 GPU. For baseline methods, we follow their official implementations. For detector-based baselines, we apply the same post-processing protocol to all models, with a confidence threshold of $\tau_{\mathrm{conf}}=0.001$ and an NMS IoU threshold of $\tau_{\mathrm{nms}}=0.25$. For IoU and MCE evaluation, each method is required to output one attention box per annotated frame. Tracking-based baselines are initialized with the ground-truth attention box in the first frame of each sequence and then evaluated on subsequent frames under the same per-frame averaging protocol. SurgLAT uses a frozen DINOv3 ViT-B/16 encoder~\cite{simeoni2025dinov3}. Input frames are resized to $512\times512$, yielding a $32\times32$ token grid, and the extracted features are projected to a hidden dimension of $C=256$. The spatial mixer uses 16 learnable evidence tokens, while the selective latent memory maintains 4 latent state tokens with short- and long-term caches of 16 and 64 previous states, respectively. Training follows a two-stage schedule: the first stage trains on short clips of length $T=32$ for 5 epochs with AdamW, a learning rate of $2\times10^{-5}$, and weight decay of $10^{-4}$; the second stage adopts online streaming training with segment length $T=64$ for another 5 epochs and a learning rate of $1\times10^{-5}$. Consecutive segments from the same video are processed chronologically, and the latent memory is carried across segment boundaries to match causal inference.

\begin{figure*}[t]
\centering
\centerline{\includegraphics[width=\linewidth]{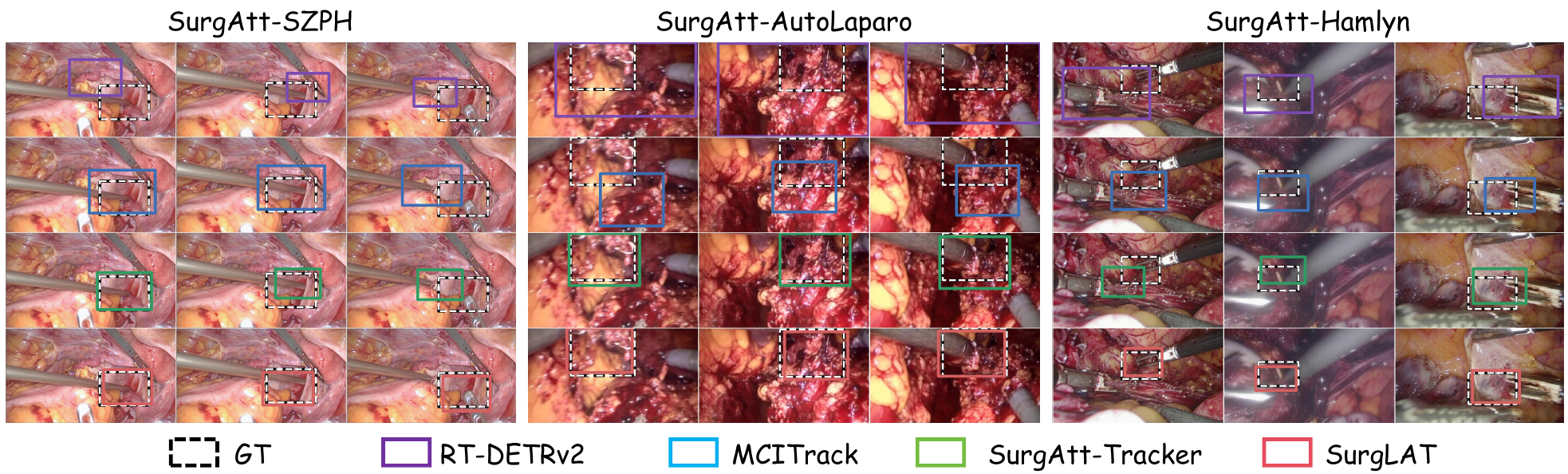}}
    \caption{
Qualitative comparison of surgical attention localization on SurgAtt-1.16M. SurgLAT shows more consistent alignment with the operative focus across in-domain and cross-domain surgical scenes.
 }
    \label{fig:qualitative_comparison}
    \vskip -0.15in
\end{figure*}

\begin{figure}[t]
\centering
\centerline{\includegraphics[width=\linewidth]{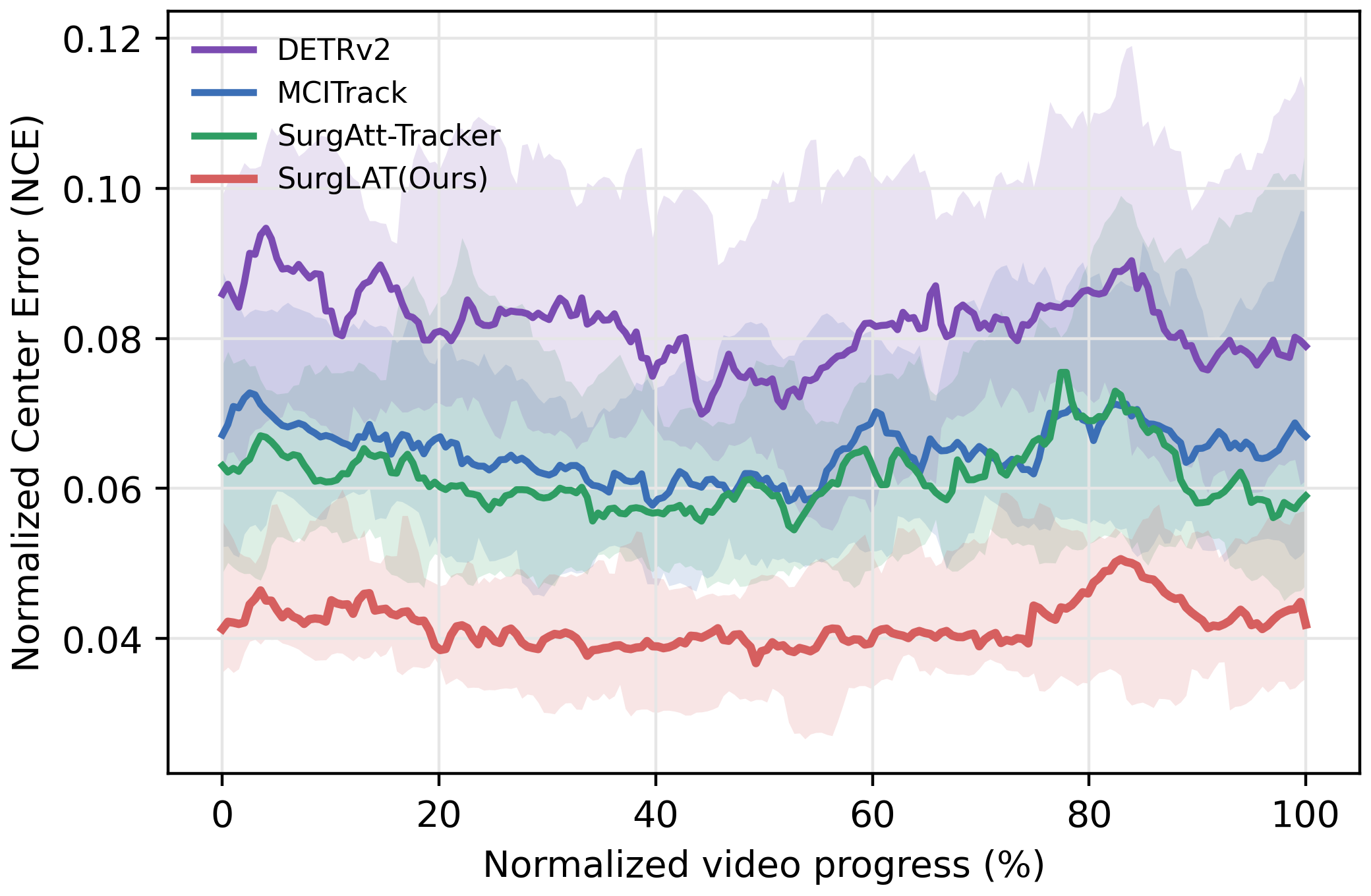}}
    \caption{
Temporal comparison of NCE across validation videos. Lower values indicate more accurate attention localization over time.}
    \label{fig:nce_curve}
    \vskip -0.05in
\end{figure}

\begin{figure}[ht]
\centering
\centerline{\includegraphics[width=\linewidth]{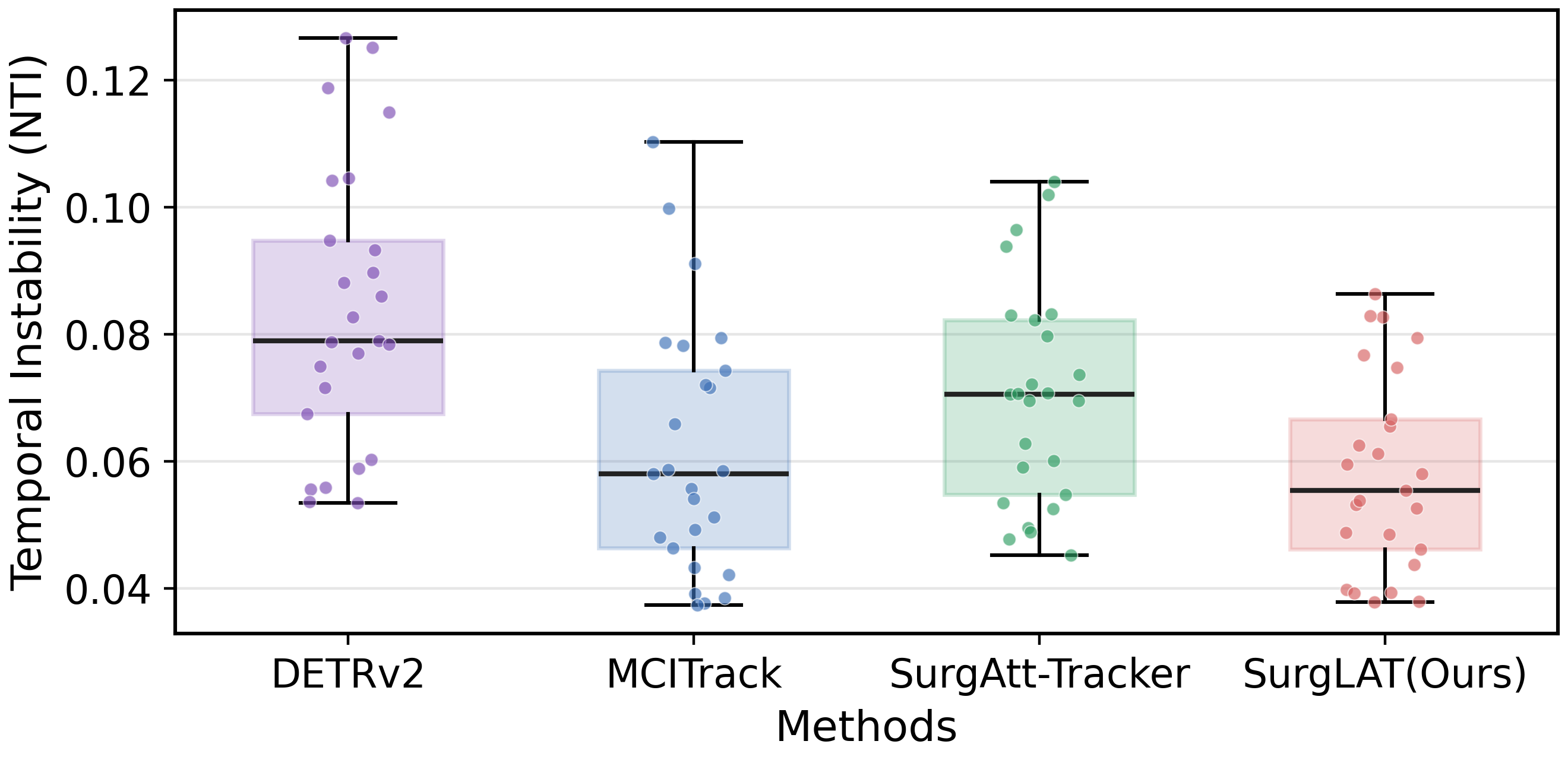}}
    \caption{
Comparison of motion-aware temporal instability across 25 validation videos. Lower values indicate smoother and more GT-consistent attention prediction.
}
    \label{fig:temporal_instability}
    \vskip -0.15in
\end{figure}

\begin{figure}[t]
\centering
\centerline{\includegraphics[width=\linewidth]{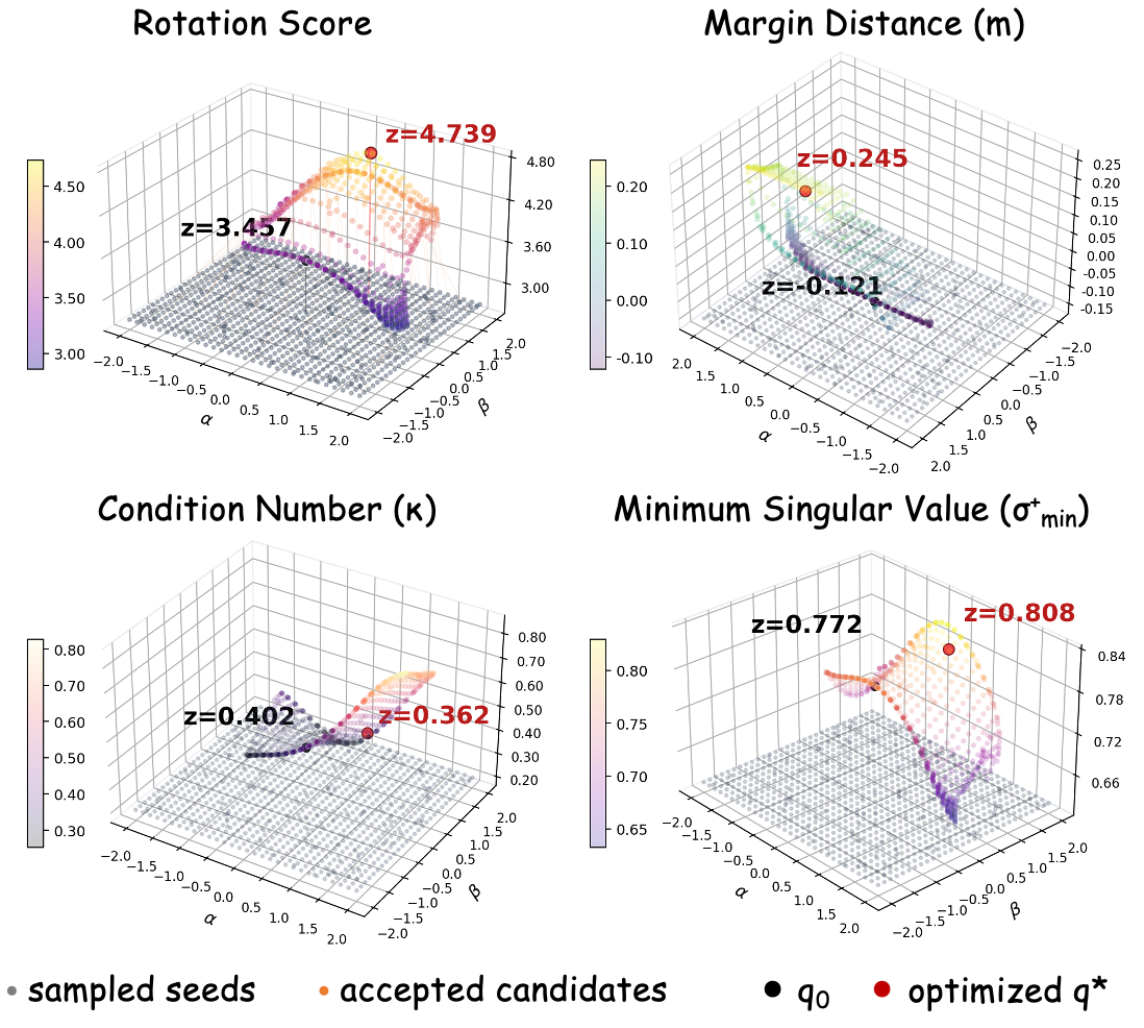}}
    \caption{
Visualization of the null-space search for initial configuration optimization. $q^{*}$ achieves improved rotational capability and a larger joint-limit margin while preserving comparable local conditioning properties.
}
    \label{fig:iniL}
    \vskip -0.1in
\end{figure}

\subsection{Quantitative Evaluation of SurgAtt-1.16M Dataset}
\subsubsection{Overall Benchmark Comparison}
Table~\ref{tab:main_results} compares SurgLAT with representative detection-based and tracking-based baselines on three surgical attention tracking benchmarks. On the in-domain SurgAtt-SZPH benchmark, SurgLAT achieves the best performance, improving IoU from 0.566 to 0.604 and reducing MCE from 49.92 to 41.24 pixels compared with the strongest baseline, SurgAtt-Tracker. The improvement is also consistent under stricter localization criteria, where mAP@0.75 increases from 0.220 to 0.268 and mAP@0.5:0.95 from 0.280 to 0.322. Under cross-domain evaluation, SurgLAT improves IoU/MCE from 0.462/121.12 to 0.527/113.75 on SurgAtt-AutoLaparo and from 0.443/42.48 to 0.479/37.97 on SurgAtt-Hamlyn. These results demonstrate more accurate operative-focus localization and better generalization across surgical domains. Fig.~\ref{fig:qualitative_comparison} further shows that SurgLAT produces attention regions more consistently aligned with the ground-truth operative focus, especially under multi-instrument interference, cluttered tissue backgrounds, and target transitions.

The performance gap reflects the limitation of treating surgical attention as either a frame-wise detection target or a conventional tracked object. Detection-based methods are efficient but lack explicit temporal state modeling, while tracking-based methods exploit temporal continuity but assume a persistent target. In laparoscopic surgery, however, the operative focus may shift among instruments, tissues, anatomical boundaries, and tool--tissue interaction regions. By modeling attention as a causal latent state, SurgLAT preserves temporal continuity during stable manipulation while adapting to attention transitions when the operative intent changes.

\subsubsection{Temporal Accuracy and Stability}
The temporal evaluation further supports the stability of SurgLAT. As shown in Fig.~\ref{fig:nce_curve}, SurgLAT maintains the lowest normalized center error over the full normalized video progress. This indicates that the proposed model provides stable online attention localization throughout the procedure, rather than achieving improvement only on isolated frames. Such temporal consistency is important for autonomous laparoscope control, where transient prediction errors may be directly converted into unnecessary camera motion. Fig.~\ref{fig:temporal_instability} compares the motion-aware temporal instability across validation videos. SurgLAT achieves the lowest and most compact instability distribution among the compared methods, demonstrating smoother and more ground-truth-consistent attention prediction. This result is particularly relevant to robotic field-of-view control, because unstable attention predictions can lead to jittery image-plane commands and oscillatory camera adjustment.

\begin{figure*}[t]
\centering
\centerline{\includegraphics[width=\linewidth]{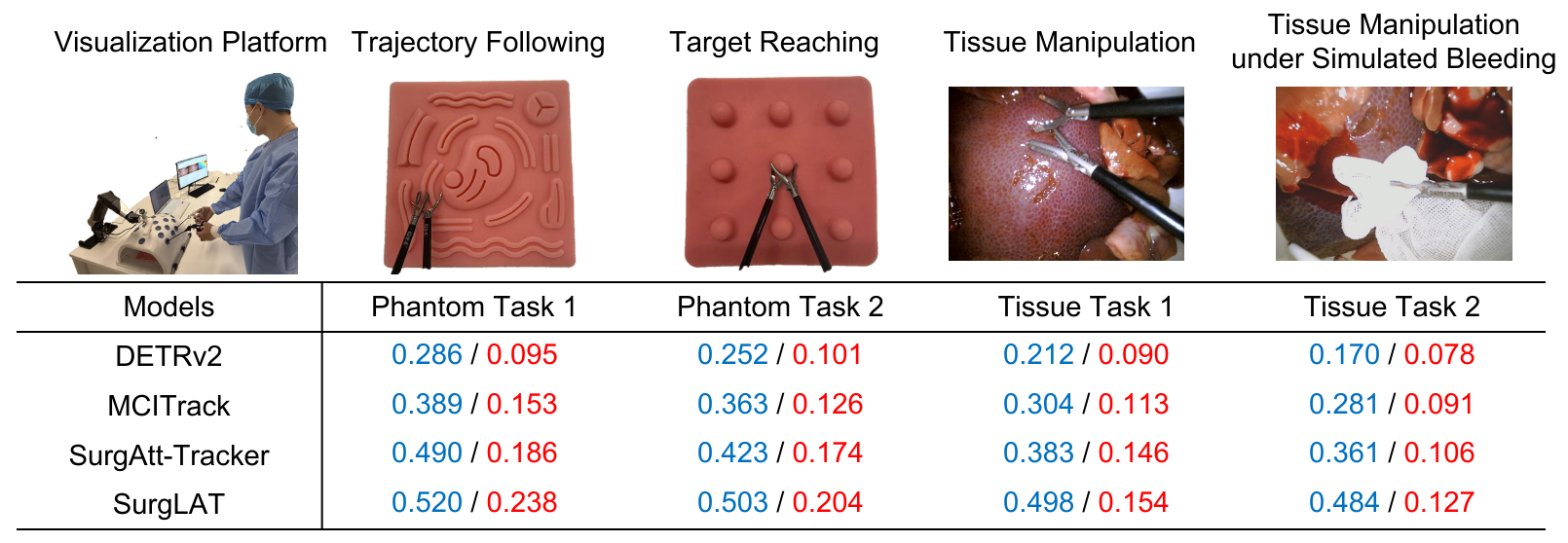}}
    \caption{
Ex vivo validation of algorithm-driven laparoscopic control across phantom and tissue manipulation tasks. Blue values denote center-reference IoU, and red values denote trajectory smoothness.}
    \label{fig:exvivo_validation}
    \vskip -0.15in
\end{figure*}

$\mathbf{a_t}=\mathbf{p}_{t+1}-2\mathbf{p}_t+\mathbf{p}_{t-1}$.

\subsection{Real-World Robotic Validation}

\subsubsection{Closed-Loop Laparoscope Control on Ex Vivo Tasks}



We further validate SurgLAT on a physical laparoscopic robotic platform under four ex vivo manipulation settings: trajectory following, target reaching, tissue manipulation, and tissue manipulation under simulated bleeding. As shown in Fig.~\ref{fig:exvivo_validation}, the predicted attention targets are converted into closed-loop camera control commands and evaluated by center-reference IoU and trajectory smoothness. SurgLAT consistently achieves the best performance across all settings. Compared with SurgAtt-Tracker, SurgLAT improves IoU/smoothness from 0.490/0.186 to 0.520/0.238 in trajectory following, from 0.423/0.174 to 0.503/0.204 in target reaching, from 0.383/0.146 to 0.498/0.154 in tissue manipulation, and from 0.361/0.106 to 0.484/0.127 under simulated bleeding. These results indicate that the proposed latent attention model provides more reliable visual targets for stable closed-loop FoV adjustment across both phantom and ex vivo tissue manipulation scenarios.

\subsubsection{RCM-Constrained Initial Configuration Optimization}
As shown in Fig.~\ref{fig:iniL}, compared with the initial configuration $q_0$, the optimized configuration $q^{*}$ increases the rotation score from $3.457$ to $4.739$ and improves the joint-limit margin from $-0.121$ to $0.245$. This indicates a larger usable rotational workspace and increased clearance from joint limits. Meanwhile, $\sigma_{\min}^{+}$ increases from $0.772$ to $0.808$, and $\kappa$ decreases from $0.402$ to $0.362$, suggesting that the optimization improves rotational feasibility with increased local kinematic conditioning. These results show that the proposed null-space search effectively exploits redundancy to obtain a more favorable initial configuration for RCM-constrained laparoscope tracking.

\begin{figure}[t]
\centering
\centerline{\includegraphics[width=\linewidth]{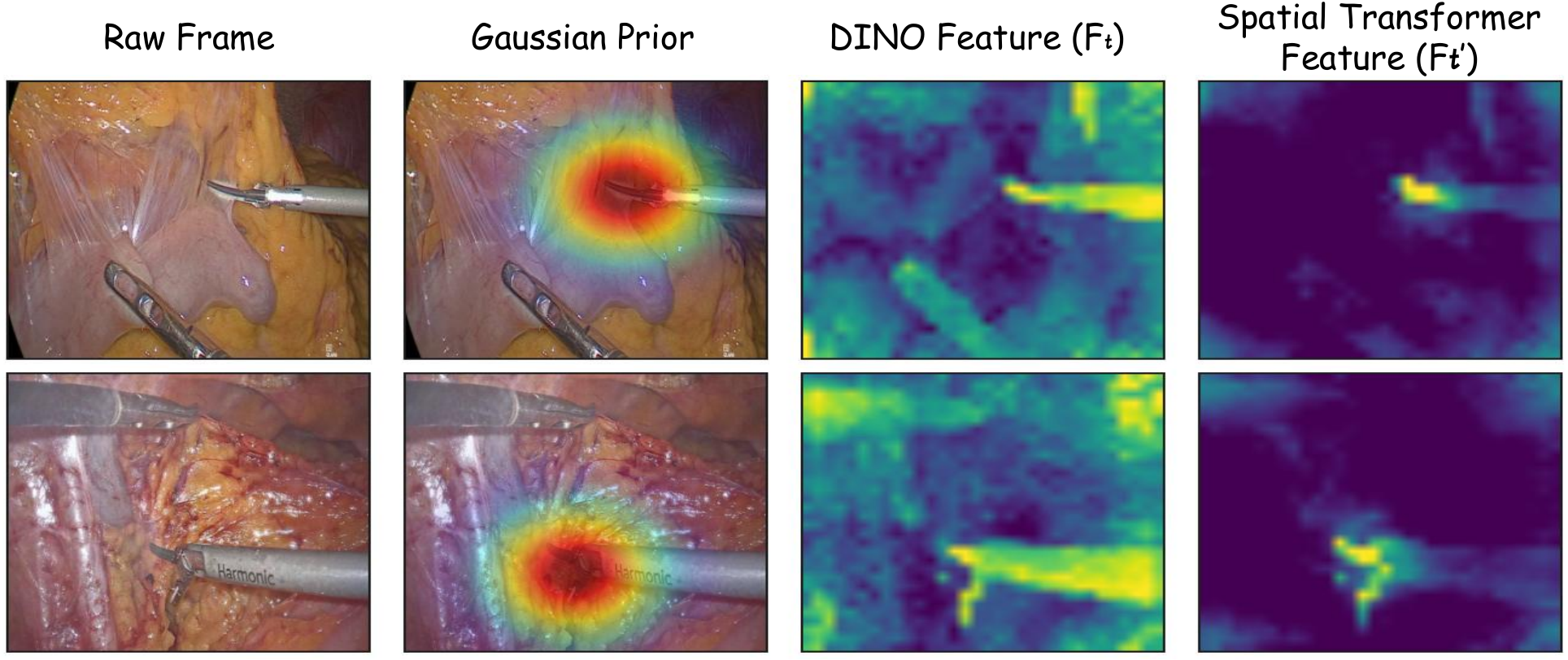}}
    \caption{
Visualization of memory-conditioned spatial reasoning. The Gaussian prior provides a coarse focus, and the spatial transformer refines broad DINO responses into a more concentrated feature activation around the operative interaction region.
}
    \label{fig:vis_feature}
    \vskip -0.15in
\end{figure}

\begin{table}[t]
\centering
\small
\caption{Ablation study of the memory-conditioned spatial prior and selective latent memory components.
}
\label{tab:ablation}

\resizebox{\linewidth}{!}{
\begin{tabular}{ccc|ccc}
\toprule

\multicolumn{3}{c|}{\textbf{Module}}
&
\multicolumn{3}{c}{\textbf{Metrics}}
\\

\midrule

{Prior Center}
&
{Short Memory}
&
{Long Memory}
&
{IoU}$\uparrow$
&
{MCE}$\downarrow$
&
{mAP@0.5}$\uparrow$
\\

\midrule

$\times$ & $\times$ & $\times$ & 0.526 & 54.87 & 0.604 \\
\checkmark & $\times$ & $\times$ & 0.571 & 48.03 & 0.638 \\
\checkmark & \checkmark & $\times$ & 0.594 & 43.72 & 0.661 \\
\checkmark & $\times$ & \checkmark & 0.586 & 45.26 & 0.653 \\
\checkmark & \checkmark & \checkmark & 0.604 & 41.24 & 0.669 \\

\bottomrule
\end{tabular}
}
    \vskip -0.15in

\end{table}

\begin{table}[t]
\centering
\small
\caption{Ablation study of different heatmap sizes and center decoding strategies.}
\label{tab:heatmap_ablation}

\resizebox{0.95\linewidth}{!}{
\begin{tabular}{ll|ccc}
\toprule

\multicolumn{2}{c|}{\textbf{Module}}
&
\multicolumn{3}{c}{\textbf{Metrics}}
\\

\midrule

{Heatmap Size}
&
{Center Decoding}
&
{IoU}$\uparrow$
&
{MCE}$\downarrow$
&
{FPS}$\uparrow$
\\

\midrule

--
& Direct center regression
& 0.548
& 53.12
& \textbf{48.6}
\\

$16\times16$
& Axis soft-argmax
& 0.581
& 46.87
& 41.2
\\

$64\times64$
& Axis soft-argmax
& \textbf{0.608}
& \textbf{40.92}
& 22.8
\\

$32\times32$
& Full soft-argmax
& 0.592
& 44.71
& 31.4
\\

$32\times32$
& Soft-argmax
& 0.598
& 43.18
& 33.1
\\

$32\times32$
& Axis soft-argmax
& \underline{0.604}
& \underline{41.24}
& \underline{34.5}
\\

\bottomrule
\end{tabular}
}
\vskip -0.15in
\end{table}

\begin{table}[t]
\centering
\small
\caption{Ablation study of heatmap, center, and temporal supervision in surgical attention localization.
}
\label{tab:loss_ablation}

\resizebox{\linewidth}{!}{
\begin{tabular}{ccc|ccc}
\toprule

\multicolumn{3}{c|}{\textbf{Loss}}
&
\multicolumn{3}{c}{\textbf{Metrics}}
\\

\midrule

$\mathcal{L}_{\mathrm{hm}}$
&
$\mathcal{L}_{\mathrm{center}}$
&
$\mathcal{L}_{\mathrm{temp}}$
&
IoU$\uparrow$
&
MCE$\downarrow$
&
mAP@0.5$\uparrow$
\\

\midrule
\checkmark & $\times$ & $\times$ & 0.575 & 48.25 & 0.634 \\
\checkmark & $\times$ & \checkmark & 0.583 & 45.72 & 0.642 \\
\checkmark & \checkmark & $\times$ & \underline{0.597} & \underline{43.90} & \underline{0.663} \\
\checkmark & \checkmark & \checkmark & \textbf{0.604} & \textbf{41.24} & \textbf{0.669} \\

\bottomrule
\end{tabular}
}
\vskip -0.15in
\end{table}

\subsection{Ablation Study}
\subsubsection{Effect of Latent Memory and Spatial Prior}

We first analyze the contribution of the memory-conditioned spatial prior and latent memory components on SurgAtt-SZPH. As shown in Table~\ref{tab:ablation}, removing all prior and memory modules leads to a clear performance drop, with an IoU of 0.526 and an MCE of 54.87 pixels. Introducing the prior center improves IoU to 0.571 and reduces MCE to 48.03, suggesting that a coarse memory-conditioned spatial cue helps guide the model toward the operative region. Adding short-term memory further improves IoU to 0.594 and MCE to 43.72 pixels, showing its importance for local motion continuity. Long-term memory also brings consistent gains, reaching 0.586 IoU and 45.26 MCE when combined with the prior center. The full model achieves the best performance, with 0.604 IoU, 41.24 MCE, and 0.669 mAP@0.5, indicating that spatial prior, short-term continuity, and long-term intent memory are complementary. Fig.~\ref{fig:vis_feature} provides a qualitative explanation of this improvement. The Gaussian prior offers a coarse focus around the expected operative region, while the DINO feature map captures broad semantic responses from instruments and tissues. After the spatial transformer, the refined feature response becomes more concentrated around the operative interaction region, supporting more accurate attention localization.

\subsubsection{Heatmap Resolution and Center Decoding}
We further evaluate the heatmap resolution and center decoding strategy in Table~\ref{tab:heatmap_ablation}. Direct center regression is computationally efficient but yields substantially lower localization accuracy, confirming that surgical attention is better represented as a spatial probability field rather than an unconstrained coordinate vector. Increasing the heatmap size from $16\times16$ to $64\times64$ improves accuracy but significantly reduces runtime. The $32\times32$ heatmap with axis soft-argmax provides the best accuracy--efficiency trade-off, achieving 0.604 IoU and 34.5 FPS. Compared with full soft-argmax, axis soft-argmax produces lower MCE while preserving real-time performance, suggesting that factorized center decoding provides a stable and efficient approximation for online attention localization.

\subsubsection{Effect of Loss Components}

Table~\ref{tab:loss_ablation} evaluates the contribution of loss components. Using only heatmap supervision achieves 0.575 IoU and 48.25 MCE, indicating that heatmap-level supervision alone is insufficient for precise center localization. Adding the center loss improves the performance to 0.597 IoU and 43.90 MCE, showing that directly constraining the decoded attention center is important for reducing localization error. When the temporal displacement loss is further introduced, the full objective achieves the best results, with 0.604 IoU, 41.24 MCE, and 0.669 mAP@0.5. These results confirm that heatmap supervision, center-level constraint, and temporal regularization provide complementary benefits for improved surgical attention prediction.

\section{Conclusion}

In this work, we presented SurgLAT, a perception-to-control framework for autonomous robotic laparoscope control. By modeling surgical attention as a causal latent state, SurgLAT integrates memory-conditioned spatial reasoning, selective short- and long-term latent memory, and heatmap-assisted ROI decoding to produce accurate and temporally stable operative targets from streaming laparoscopic video. The predicted attention target is further combined with depth-aware operative scale estimation and executed through a virtual-axis-based RCM-constrained controller with redundancy-aware null-space initialization. Experiments on SurgAtt-1.16M demonstrate improved attention localization, temporal stability, and cross-domain generalization over detection- and tracking-based baselines, while ex vivo robotic validation shows stable closed-loop FoV adjustment across phantom and tissue manipulation tasks. Although the current robotic experiments are conducted mainly in phantom and ex vivo settings, future work will extend SurgLAT toward larger-scale in vivo validation, more complex surgical workflows, and stronger safety-aware control constraints for clinical deployment.

\bibliographystyle{IEEEtran}
\bibliography{references}

\end{document}